\newif\ifarxiv

\arxivtrue

\ifarxiv
    \documentclass{article}
    \usepackage{arxiv}
    \usepackage[utf8]{inputenc} 
    \usepackage[T1]{fontenc}    
    \usepackage{hyperref}       
    \usepackage{url}            
    \usepackage{booktabs}       
    \usepackage{amsfonts}       
    \usepackage{nicefrac}       
    \usepackage{microtype}      
    \usepackage{amsmath}

    \usepackage{lipsum}
    \usepackage{graphicx}
    \usepackage[norule,symbol,perpage]{footmisc}

        \usepackage{caption}
    \usepackage{subcaption}
        \usepackage{bm} 
      \graphicspath{{imgs/}}
\else
\documentclass[conference,a4paper,twocolumn]{IEEEtran}
\IEEEoverridecommandlockouts

\usepackage{tipa}
\usepackage[pdftex]{graphicx}
\usepackage{epsfig}
\usepackage{epstopdf}
\usepackage[cmex10]{amsmath}
\usepackage{stfloats}
\usepackage{url}

\usepackage{amsmath}
\usepackage{booktabs}
\usepackage{caption}
\usepackage{subcaption}
\usepackage{multirow}
\usepackage{soul,color}
\usepackage[utf8x]{inputenc}
\usepackage{etoolbox}
\usepackage{amssymb}
\usepackage{breqn}
   \usepackage[pdftex]{graphicx}
   \graphicspath{{imgs/}}
   \DeclareGraphicsExtensions{.pdf,.jpeg,.png}

\usepackage{mathtools}

\fi

\begin{document}

\title{Comparing Normalization Methods for Limited Batch Size Segmentation Neural Networks}


\ifarxiv
\author{
    Martin Kolarik, Radim Burget, Kamil Riha\\
Dept. of Telecommunications\\
Brno University of Technology\\
Brno, Czech Republic\\
\texttt{martin.kolarik@vutbr.cz}}
\fi


%


\maketitle

\begin{abstract}
The widespread use of Batch Normalization has enabled training deeper neural networks with more stable and faster results. However, the Batch Normalization works best using large batch size during training and as the state-of-the-art segmentation convolutional neural network architectures are very memory demanding, large batch size is often impossible to achieve on current hardware. We evaluate the alternative normalization methods proposed to solve this issue on a problem of binary spine segmentation from 3D CT scan. \\
Our results show the effectiveness of Instance Normalization in the limited batch size neural network training environment. Out of all the compared methods the Instance Normalization achieved the highest result with Dice coefficient = 0.96 which is comparable to our previous results achieved by deeper network with longer training time. We also show that the Instance Normalization implementation used in this experiment is computational time efficient when compared to the network without any normalization method.
\end{abstract}




\section{Introduction}
\footnotemark[0]Batch Normalization (BatchNorm) introduced by Ioffe and Szegedy at 2015 \cite{batchnorm} has quickly become a widely used method for accelerating and stabilizing the training process of deep convolutional neural networks (CNNs). The core idea behind the need for normalization within the network is to suppress the effect of internal covariate shift which leads to excessive saturation of CNN weights. Newer works also point that the usage of BatchNorm leads to a smoother optimization landscape and therefore better and faster convergence to a global minimum \cite{santurkar2018does} \cite{bjorck2018understanding}. \footnotetext[0]{This paper has been accepted for publishing at the TSP 2020 Conference}\\
The nature of the Batch Normalization algorithm requires large batch size (many batches processed in parallel) during training process in order to obtain statistically significant mean and standard deviation (std) of the input data. This is a limiting factor for the usage of BatchNorm especially in deep 3D segmentation CNNs where the algorithm is constrained by available GPU memory. It is not uncommon to train such networks using batch size only equal to one. This is the reason for development of new CNN normalization techniques which are independent of batch size hyperparameter. This paper aims to compare the BatchNorm with Group normalization (GroupNorm) \cite{groupnorm} and Instance normalization (InstanceNorm) \cite{instancenorm} in terms of computational time, efficiency and resulting accuracy for 3D segmentation CNNs.

\begin{figure}[!t]
\centering
     \begin{subfigure}[b]{0.51\linewidth}
         \centering
         \includegraphics[width=0.7\linewidth]{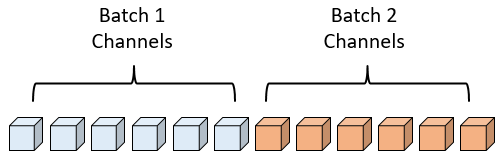}
         \caption{Visualisation of Batch Normalization input data set $S_i$}
         \label{fig:batchnorm}
     \end{subfigure}
          \begin{subfigure}[b]{0.51\linewidth}
         \centering
         \includegraphics[width=0.7\linewidth]{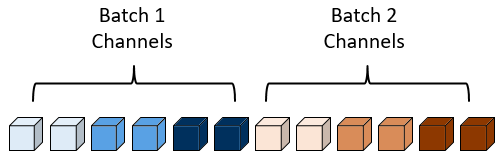}
         \caption{Visualisation of Group Normalization with 6 groups data set $S_i$}
         \label{fig:groupnorm}
     \end{subfigure}
          \begin{subfigure}[b]{0.51\linewidth}
         \centering
         \includegraphics[width=0.7\linewidth]{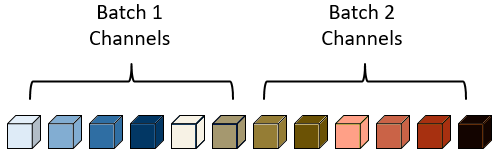}
         \caption{Visualisation of Instance Normalization input data set $S_i$}
         \label{fig:instancenorm}
     \end{subfigure}
        \caption{Visualisation of compared normalization methods data input distribution - same color mark groups of the input data set $S_i$ in which the mean and std is computed.}
        \label{fig:normalizationmethods}
        \vspace{-20pt}
\end{figure}

\section{Related works}
General formula of feature normalization can be seen in Eq.~\ref{eqgeneralnormalization} where \textit{x} is the feature output of a layer and \textit{i} is an index of \textit{x}. Different types of normalization vary in the definition of \textit{x} and chosen type of indexing \textit{i}. In case of 3D images, \textit{x} is a 5D vector (N, D, H, W, C) where N stands for batch axis, D is feature depth, H height and W width respectively. Index C stands for channel dimension and the $i_C$ is the usual choice for normalization indexing out of possible \textit{i = ($i_N$, $i_D$, $i_H$, $i_W$, $i_C$)}

\begin{equation}
\label{eqgeneralnormalization}
\hat{x}_i = \frac{1}{\sigma_i}(x_i - \mu_i)
\end{equation}

Following Eq.~\ref{eqgeneralmeanstd} shows the formula for calculating the mean $\mu$ and the standard deviation $\sigma$.

\begin{equation}
\label{eqgeneralmeanstd}
\mu_i = \frac{1}{m}\sum_{k\in S_i} x_k,\;\;\;  \sigma_i = \sqrt{\frac{1}{m}\sum_{k\in S_i}(x_k - \mu_i) + \epsilon}
\end{equation}

The $\epsilon$ in Eq.~\ref{eqgeneralmeanstd} stands for a small constant. The important variable in Eq.~\ref{eqgeneralmeanstd} is the set of features $S_i$ in which the mean and std are computed. Discussed types of normalization differ in the definition of $S_i$ as can be seen in Fig.~\ref{fig:normalizationmethods} \cite{groupnorm}.
Generally speaking the difference between the BatchNorm, the GroupNorm and the InstanceNorm lies in the granularity into which you divide your input set of features $S_i$. The BatchNorm calculates the $\mu$ and $\sigma$ over each batch (corresponding channels in each batch), the GroupNorm calculates the $\mu$ and $\sigma$ over chosen number of groups instead of batches and the InstanceNorm calculates the $\mu$ and $\sigma$ over corresponding channels (each sample of each channel) without further grouping. 

\subsection{Batch normalization}
Batch Normalization is the most frequently used method of feature normalization. We can define the BatchNorm operation as Eq.~\ref{eqsetbatchnorm} where $i_C$ (and $k_C$) states the sub-index of i (and k respectively) along the channel axis C. The BatchNorm computes $\mu$ and $\sigma$ along the (N, D, H, W) axes \cite{groupnorm}.

\begin{equation}
\label{eqsetbatchnorm}
S_i =  \{k\;|\;k_C = i_C\}
\end{equation}

To illustrate the problem of BatchNorm for small batch sizes see Eq.~\ref{eqbatchmean}-\ref{eqbatchscaleshift} \cite{batchnorm}. As can be seen in Eq.~\ref{eqbatchmean} and Eq.~\ref{eqbatchvariance}, using BatchNorm layers together with batch size one results in zero variance therefore the denominator in Eq.~\ref{eqbatchnormalization} becomes very large. The otherwise beneficial BatchNorm starts to add error to the training process due to wrong batch statistics estimation and also the learnable parameters $\gamma$ and $\beta$ will take wrong values.

\begin{equation}
\label{eqbatchmean}
\mu_\beta \xleftarrow[]{} \frac{1}{m} \sum_{i=1}^{m} x_i
\end{equation}

\begin{equation}
\label{eqbatchvariance}
\sigma^2_\beta \xleftarrow[]{} \frac{1}{m} \sum_{i=1}^{m}(x_i - \mu_\beta)^2
\end{equation}

\begin{equation}
\label{eqbatchnormalization}
\hat{x}_i \xleftarrow[]{} \frac{x_i - \mu_\beta}{\sqrt{\sigma^2_\beta + \epsilon}}
\end{equation}

\begin{equation}
\label{eqbatchscaleshift}
y_i \xleftarrow[]{} \gamma \hat{x}_i + \beta \equiv \textbf{BN}_{\gamma, \beta}(x_i)
\end{equation}

\subsection{Group normalization}
One of the proposed solutions to the BatchNorm small batch size problem is the Group Normalization (GroupNorm) \cite{groupnorm}. GroupNorm layer computes the $\mu$ and $\sigma$ in a set $S_i$ defined in Eq.~\ref{eqsetGroupnorm}.

\begin{equation}
\label{eqsetGroupnorm}
S_i =  \{k\;|\;k_N = i_N, \lfloor\frac{k_C}{C/G}\rfloor = \lfloor\frac{i_C}{C/G}\rfloor\}
\end{equation}

In Eq.~\ref{eqsetGroupnorm} the G is a hyperparameter setting number of groups into which the channels are divided. The GroupNorm computes $\mu$ and $\sigma$ along the 3D vector (D,H,W) axes along the group of $\frac{C}{G}$ channels \cite{groupnorm}.

\subsection{Instance normalization}
Instance normalization (InstanceNorm) has been proposed as a method for style transfer CNNs \cite{instancenorm}. It can be successfully used as a drop-in replacement for BatchNorm. For 3D networks the InstanceNorm computes the $\mu$ and $\sigma$ along the (D,H,W,C) axes, the set $S_i$ defined in Eq.~\ref{eqsetInstancenorm}.

\begin{equation}
\label{eqsetInstancenorm}
S_i =  \{k\;|\;k_N = i_N,k_C = i_C\}
\end{equation}

Note that the GroupNorm is the same as InstanceNorm in case we choose the parameter G = \textit{number of input channels}.

\subsection{Other normalization methods}
Family of normalization methods includes also notable approaches of Layer normalization \cite{layernorm} and Weight Normalization \cite{weightnorm}. As these have been primarily developed for other types of neural networks such as recurrent networks, we did not include them in the comparison experiment and their detailed explanation is above the scope of this paper.



\section{Methodology}
The experiment is designed to evaluate the properties of each normalization algorithm in terms of computational demands, training time and the final reached accuracy measured in Dice coefficient. We have evaluated the algorithms using the batch size one in order to test their behavior in limited batch size environment due to the current deep segmentation CNNs memory demands.

\subsection{Dataset}
The dataset used in this experiment is a CT transversal thoracic spine segmentation dataset. The original dataset was used in the segmentation challenge of the CSI Miccai 2014 Workshop \cite{yao2012detection}. We modified the data so it would be suitable for the problem of binary semantic segmentation and instead of identifying each vertebra with different mask value, we set all voxels belonging to spine to a value of 255 and the rest to zero. The spine dataset consists of 10 healthy subjects in the age between 16-35 years. Scans cover lumbar and thoracic spine region. Data was provided in the NRRD format. Example of an input scan is at Fig.~\ref{fig:comparison_scan} and the reference mask is in Fig.~\ref{fig:comparison_referencemask}.

\subsection{Data preparation}
The input data was resampled into slice resolution of 256 $\times$ 256 and scaled between interval [0,1] for the neural network input. The groundtruth mask were normalized also between [0,1] because the network uses the sigmoid activation convolutional layer as output. We divided the input data scan slices into overlapping batches of 16 slices each resulting in input and output 3D feature size 16 $\times$ 256 $\times$ 256. The batch was sharing first 8 slices with the preceding batch and the last 8 slices with the subsequent one. We exploited this overlapping property during prediction phase when we composed the resulting output extracting only the middle 8 slices from each batch. The output was scaled into interval [0,255] and thresholded with a threshold value 128 to obtain final segmentation.

\subsection{Neural network architecture}
\begin{figure*}[!t]
\centering
\subfloat{\includegraphics[width=0.95\textwidth]{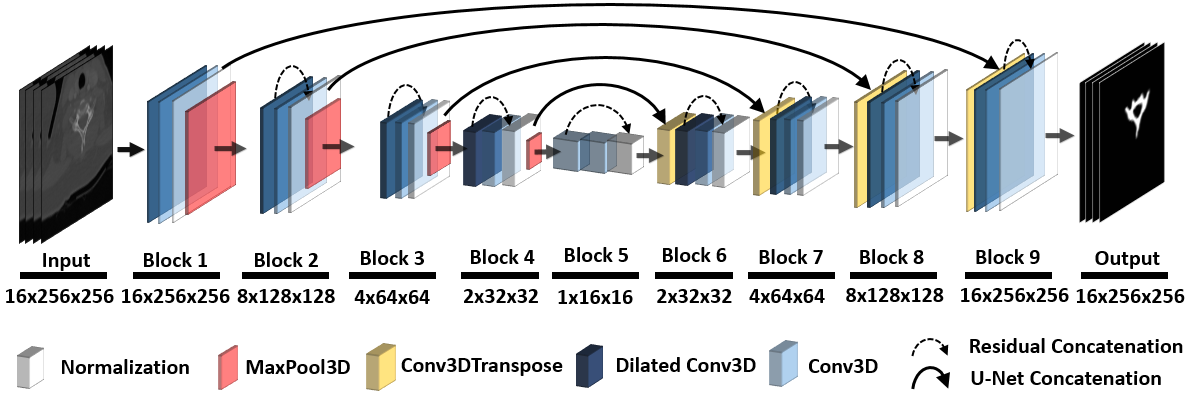}}
\captionsetup{justification=centering}
\caption{Architecture of the 3D res-u-net network used in this experiment. \label{fig:architecture}}
        \vspace{-20pt}
\end{figure*}

The neural network implementation is inspired by our previous research \cite{kolavrik2019optimized} and can be seen in Fig.~\ref{fig:architecture}. The used architecture 3D res-u-net is based on auto-encoder U-Net network \cite{ronneberger2015u} and leverages the residual concatenations. Residual blocks consist of dilated \cite{yu2015multi} and standard Conv3D with following normalization layer. \\
Using the feature size of 16 $\times$ 256 $\times$ 256 the network was able to fit with batch size 2 during training on rtx 2080ti GPU. For the purpose of this experiment we tested all networks with batch size 1 and all following results will be measured with batch size 1. This is in order to simulate environment with deeper network and larger feature size.

\subsection{Implementation details}
The experiment is implemented in Keras \cite{chollet2015keras} with Tensorflow backend \cite{tensorflow2015-whitepaper}. The hardware used for computation is a Nvidia rtx 2080ti GPU with 11 gb of GPU memory. The performance of the measured normalization method was based on the achieved Dice coefficient at Eq.~\ref{eqDice}.

\begin{equation}
\label{eqDice}
D(X,Y) = \frac{2*\left|X \cap Y\right|}{\left|X\right|+\left|Y\right|} = \frac{2 * TP}{\left|X\right|+\left|Y\right|}
\end{equation}

We chose the combination of binary crossentropy (BC) and dice coefficient as loss function, see subsequent equation Eq.~\ref{eqLoss} for more details of the used loss.

\begin{equation}
\label{eqLoss}
Loss(X,Y) = BC(X,Y) - D(X,Y) + 1
\end{equation}

All networks were trained for 30 epochs with a learning rate $5e-5$. This combination resulted in divergence during training of the network without normalization. This further shows the importance of normalization method for training stabilization. \\
The used implementation of BatchNorm is the default Keras Batchnormalization layer. GroupNorm implementation has been used from the repository \cite{groupnormgithub} and InstanceNorm from the Keras-contrib branch \cite{instancenormgithub}.

\section{Results}

We evaluated the 3D res-u-net network performance with BatchNorm, GroupNorm with parameter G = (2,4,8,16,32), InstanceNorm and for comparison also without any normalization method. Results of the segmentation network with each implemented normalization method can be seen in Tab.~\ref{tab:time_results} and Tab.~\ref{tab:final_results}.

\begin{table}[!h]
\captionsetup{labelfont=bf}
\renewcommand{\arraystretch}{1.3}
\caption{Results - comparison of the measured normalization method based on the training and prediction time requirements}
\label{tab:time_results}
\centering
\begin{tabular}{lcc}
\toprule
        Method [groups]  & Time per epoch [s] & Prediction time [s]\\
\midrule
         Without normalization&    131&    8 \\
         Batch normalization&    152&    11 \\
         Group normalization [G=2]&    228&    12 \\
         Group normalization [G=4]&    223&    11 \\
         Group normalization [G=8]&    222&    11 \\
         Group normalization [G=16]&   222&    11 \\
         Group normalization [G=32]&    202&    11 \\
         Instance normalization&    163&    8 \\
         
\bottomrule
\end{tabular}
\end{table}

\begin{table}[!h]
\captionsetup{labelfont=bf}
\renewcommand{\arraystretch}{1.3}
\caption{Results - comparison of the measured normalization method based on the achieved Dice coefficient metric}
\label{tab:final_results}
\centering
\begin{tabular}{lcc}
\toprule
        Method  & Group size&         Dice coef. \\
\midrule
         Without normalization&    -&    0.598 \\
         Batch normalization&    -&    0.888 \\
         Group normalization&    2&    0.941 \\
         Group normalization&    4&    0.926 \\
         Group normalization&    8&    0.918 \\
         Group normalization&    16&    0.920 \\
         Group normalization&    32&    0.916 \\
         Instance normalization&    -&    \textbf{0.960} \\
         
\bottomrule
\end{tabular}
\end{table}

\begin{figure*}[!t]
\captionsetup{justification=centering}
\centering
     \begin{subfigure}[b]{0.16\textwidth}
         \centering
         \includegraphics[width=1\textwidth]{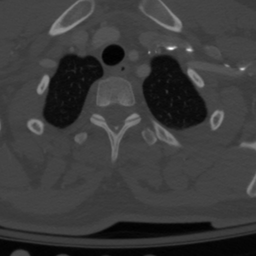}
         \caption{CT input scan}
         \label{fig:comparison_scan}
     \end{subfigure}
     \begin{subfigure}[b]{0.16\textwidth}
         \centering
         \includegraphics[width=1\textwidth]{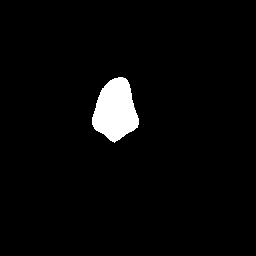}
         \caption{Without norm.}
         \label{fig:comparison_without}
     \end{subfigure}
     \begin{subfigure}[b]{0.16\textwidth}
         \centering
         \includegraphics[width=1\textwidth]{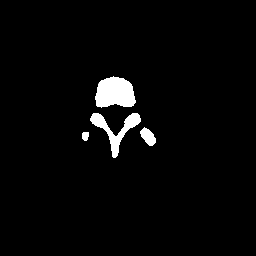}
         \caption{BatchNorm}
         \label{fig:comparison_batchnorm}
     \end{subfigure}
     \begin{subfigure}[b]{0.16\textwidth}
         \centering
         \includegraphics[width=1\textwidth]{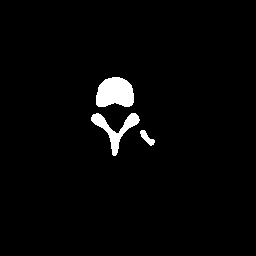}
         \caption{GroupN. [G=2]}
         \label{fig:comparison_groupnorm}
     \end{subfigure}
     \begin{subfigure}[b]{0.16\textwidth}
         \centering
         \includegraphics[width=1\textwidth]{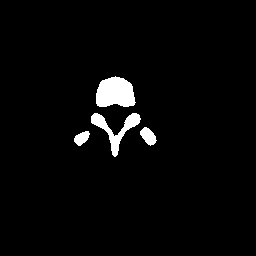}
         \caption{InstanceNorm}
         \label{fig:comparison_instancenorm}
     \end{subfigure}
     \begin{subfigure}[b]{0.16\textwidth}
         \centering
         \includegraphics[width=1\textwidth]{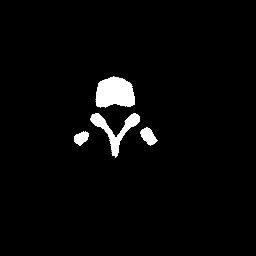}
         \caption{Reference mask}
         \label{fig:comparison_referencemask}
     \end{subfigure}
     \captionsetup{justification=centering}
        \caption{Examples of (from left) the input CT transversal scan, prediction of CNN without normalization, prediction using BatchNorm, prediction using GroupNorm, prediction using InstanceNorm and the reference groundtruth mask.}
        \label{fig:comparison}
\end{figure*}


\section{Discussion}
The InstanceNorm has achieved best result in comparison with other methods when measured in Dice coefficient and its implementation shows promising computational time requirements. The GroupNorm longer training times are a result of different implementation, but the method did not achieve as good results as the InstanceNorm. Predicted masks using the compared methods are in Fig.~\ref{fig:comparison}.\\
As expected, the BatchNorm with batch size = 1 did not achieve good results. Surprisingly the network without normalization did not converge to a good solution partly due to the BCeDice loss function. This clearly shows the need for feature normalization.

\section{Conclusion}
In this paper we have compared different normalization methods with focus on approaches suitable for limited batch size deep segmentation CNNs. Our result show that for small batch size the InstanceNorm achieved best result with Dice coefficient = 0.96 which is comparable to our previous results on the same dataset with more complex network \cite{kolavrik2019optimized}. \\
We plan to incorporate the Instance Normalization in our future experiments with deep 3D CNN transfer learning to achieve higher accuracy on medical 3D limited datasets.


\section*{Acknowledgment}
Research described in this paper was supported by the MPO FV20044, National Sustainability Program under grant LO1401 and by European Regional Development Fund, project Interreg, niCE-life, CE1581. Supported by Ministry of Health of the Czech Republic, grant nr. NV18-08-00459. All rights reserved.

\newpage
\bibliographystyle{IEEEtran}
\bibliography{bibliography.bib}




\end{document}